\documentclass[a4paper]{article}
\usepackage[english]{babel}
\usepackage{INTERSPEECH2021}
\usepackage{xcolor}

\title{Transducer-based language embedding for spoken language identification}
\name{Peng Shen, Xugang Lu, Hisashi Kawai}
\address{
  National Institute of Information and Communications Technology (NICT)
}
\email{peng.shen@nict.go.jp}

\begin{document}

\maketitle
\begin{abstract}
  The acoustic and linguistic features are important cues for the spoken language identification (LID) task. Recent advanced LID systems mainly use acoustic features that lack the usage of explicit linguistic feature encoding.
  In this paper, we propose a novel transducer-based language embedding approach for LID tasks by integrating an RNN transducer model into a language embedding framework.
  Benefiting from the advantages of the RNN transducer's linguistic representation capability, the proposed method can exploit both phonetically-aware acoustic features and explicit linguistic features for LID tasks.
  Experiments were carried out on the large-scale multilingual LibriSpeech and VoxLingua107 datasets.
  Experimental results showed the proposed method significantly improves the performance on LID tasks with 12\% to 59\% and 16\% to 24\% relative improvement on in-domain and cross-domain datasets, respectively.
\end{abstract}

\section{Introduction}
Spoken language identification (LID) is one of the most important steps in multi-lingual speech processing systems. Recently, with the rapid development of deep learning algorithms, LID technologies have also made rapid progress \cite{DehakIEEE2015, LOPEZMORENO201646, shen2016, LozanoDiez2015, Fernando2017, GengWang2016,shen2018feature}.
For LID tasks, not only acoustic features, such as phonotactics information but also linguistic features, such as contextual information, are important cues to determine a language \cite{HaizhouLi2013}.
The recent deep neural network-based LID technologies significantly improve the accuracy of LID by using a large amount of training data and complex network structure with powerful acoustic feature extraction and abstraction capability.
However, in real applications, such techniques often suffer from overfitting problems because the recording conditions and speaking styles of a test dataset are different from those of the training dataset, i.e., the cross-domain problem.

To further improve the performance and overcome the overfitting problem of LID systems, previous studies have proposed to use the phonetic information and contextual information for building LID systems \cite{lizheng2021additivePhoneme, Ren2019TwostageTF,jicheng2021E2EmultiTask, LiBo2018multiDialectASR}.
Most of these works use speech recognition models to enhance the LID system.
One approach is fine-tuning a pre-trained automatic speech recognition (ASR) model for LID tasks. For example, Ren et al. proposed a two-stage training strategy in which an acoustic model firstly is trained with connectionist temporal classification (CTC) to recognize the given phonetic sequence annotation, then a recurrent neural network is used to classify language category by utilizing the intermediate features as inputs from the acoustic model \cite{Ren2019TwostageTF}.
Another approach is the multi-task training method.
Such methods improve performance and model robustness by sharing the underlying feature extraction network and jointly training objective functions of speech/phoneme recognition and language recognition \cite{lizheng2021additivePhoneme, jicheng2021E2EmultiTask, LiBo2018multiDialectASR}.
Recently, self-supervised phonotactic representations were also investigated on LID tasks \cite{ramesh21_interspeech, Fanzhiyun2021wav2vec4LID}.
The methods described above are based on the acoustic module of a speech recognition model or the implicit linguistic expression of an end-to-end (E2E) ASR model. Therefore they lack the usage of the explicit linguistic expression of the traditional language model.

In this paper, we propose a novel transducer-based language embedding approach by integrating an RNN transducer (RNN-T) model into a language embedding extraction framework.
Recently, RNN-T becomes the most popular E2E ASR model for its natural streaming capability and impressive performance \cite{Alex2012RNNT, jinyuLi2021E2EASR}.
Different from the conventional CTC-based E2E model, the RNN-T includes not only an encoder network for acoustic feature extraction but also a prediction network for explicit linguistic feature extraction.
Benefiting from the advantages of the RNN-T's linguistic representation capability, the proposed method can exploit both phonetically-aware acoustic features and explicit linguistic features for LID tasks.
Different from the previous work investigating RNN-T with LID mainly focus on dynamic languages switching for speech recognition task \cite{amazon2021jontASRLID}, Our work is the first to focus on integrating RNN-T into language embedding extraction for LID tasks.
The contributions of this work can be summarized as follows:
\begin{itemize}
  \item We propose a novel RNN-T-based language embedding extraction method by using both the acoustic and explicit linguistic features.
  \item Early additive fusion and late dual-branch statistic pooling fusion methods are investigated on integrating the acoustic and linguistic representation of RNN-T.
  \item We analyze the characteristics of RNN-T and propose a correlation alignment to align the outputs of encoder and prediction networks. Experiments are designed to illustrate how the RNN-T modules can be better applied for language embedding extraction.
\end{itemize}
We evaluated the proposed method on the multilingual LibriSpeech (MLS) and VoxLingua107 datasets, the experimental results showed that the proposed method could significantly improve the LID performance not only on the in-domain dataset but also on the cross-domain dataset.

\section{RNN Transducer}
RNN-T becomes the most popular E2E model in the industry for it provides a natural way for streaming ASR systems  \cite{Alex2012RNNT, jinyuLi2021E2EASR}.
Compared with the conventional CTC, RNN-T removes the conditional independence assumption, it outputs conditions on the previous output tokens and the speech sequence until the current time step as $P(y_u|\mathbf{x}_{1:t},\mathbf{y}_{1:u})$.
An RNN-T consists of an encoder network, a prediction network, and a joint network.
The encoder network plays an acoustic feature extraction role by generating a high-level feature representation $\mathbf{h}^{enc}_t$.
The prediction network is a language model that produces a linguistic representation $\mathbf{h}^{pre}_u$ based on RNN-T's previous output label $y_{u-1}$.
Then $\mathbf{h}^{enc}_t$ and $\mathbf{h}^{pre}_u$ are combined with the feed-forward network-based joint network as
\begin{equation}\label{eq.rnnt1}
  \mathbf{z}_{t,u} =  \psi(\mathbf{Q} \mathbf{h}^{enc}_t + \mathbf{V} \mathbf{h}^{pre}_u + \mathbf{b}_z),
\end{equation}
where $\psi$ is a non-linear function, $\mathbf{Q}$ and $\mathbf{V}$ denote weight matrices, and $\mathbf{b}_z$ is a bias vector. Then the output $\mathbf{h}_{t,u}$ of the joint network is obtained by applying a linear transform on the input of $\mathbf{z}_{t,u}$ as
\begin{equation}\label{eq.rnnt2}
  \mathbf{h}_{t,u} = \mathbf{W}_y \mathbf{z}_{t,u} + \mathbf{b}_y,
\end{equation}
where $\mathbf{W}_y$ is a weight matrix, and $\mathbf{b}_y$ denote a bias vector.
Finally, the probability of each output symbol $k$ is defined as
\begin{equation}\label{eq.rnn3}
  P(y_u = k|\mathbf{x}_{1:t}, \mathbf{y}_{1:u-1}) = softmax(\mathbf{h}^k_{t,u}).
\end{equation}
The objective function of RNN-T is defined as $- ln P(\mathbf{y}|\mathbf{x})$, where $P(\mathbf{y}|\mathbf{x})$ is the sum of all possible alignment paths that are mapped to the label sequence which can be described as
\begin{equation}\label{eq.rnn4}
  P(\mathbf{y}|\mathbf{x}) = \sum_{\mathbf{a} \in A^{-1}(\mathbf{y})} P(\mathbf{a}|\mathbf{x}),
\end{equation}
where $\mathbf{a}$ is alignment paths. The mapping from the alignment
path $\mathbf{a}$ to the label sequence $\mathbf{y}$ is defined as $A(\mathbf{a}) = \mathbf{y}$.

\section{Transducer-based language embedding}
Our LID system consists of a front-end feature extraction and a back-end classifier. The front-end and back-end-based framework was widely used in both LID and speaker recognition tasks \cite{snyder2017xvector, Li2017DeepSA, snyer2018xvector4LID}. It obtains state-of-the-art performance for LID challenge tasks \cite{AP20}.
In this section, we first present the implementation of the proposed front-end transducer-based language embedding method.
Then, the back-end classifier for language recognition is introduced.

\subsection{Front-end architecture}
\begin{figure}
  \centering
  \includegraphics[width=230px]{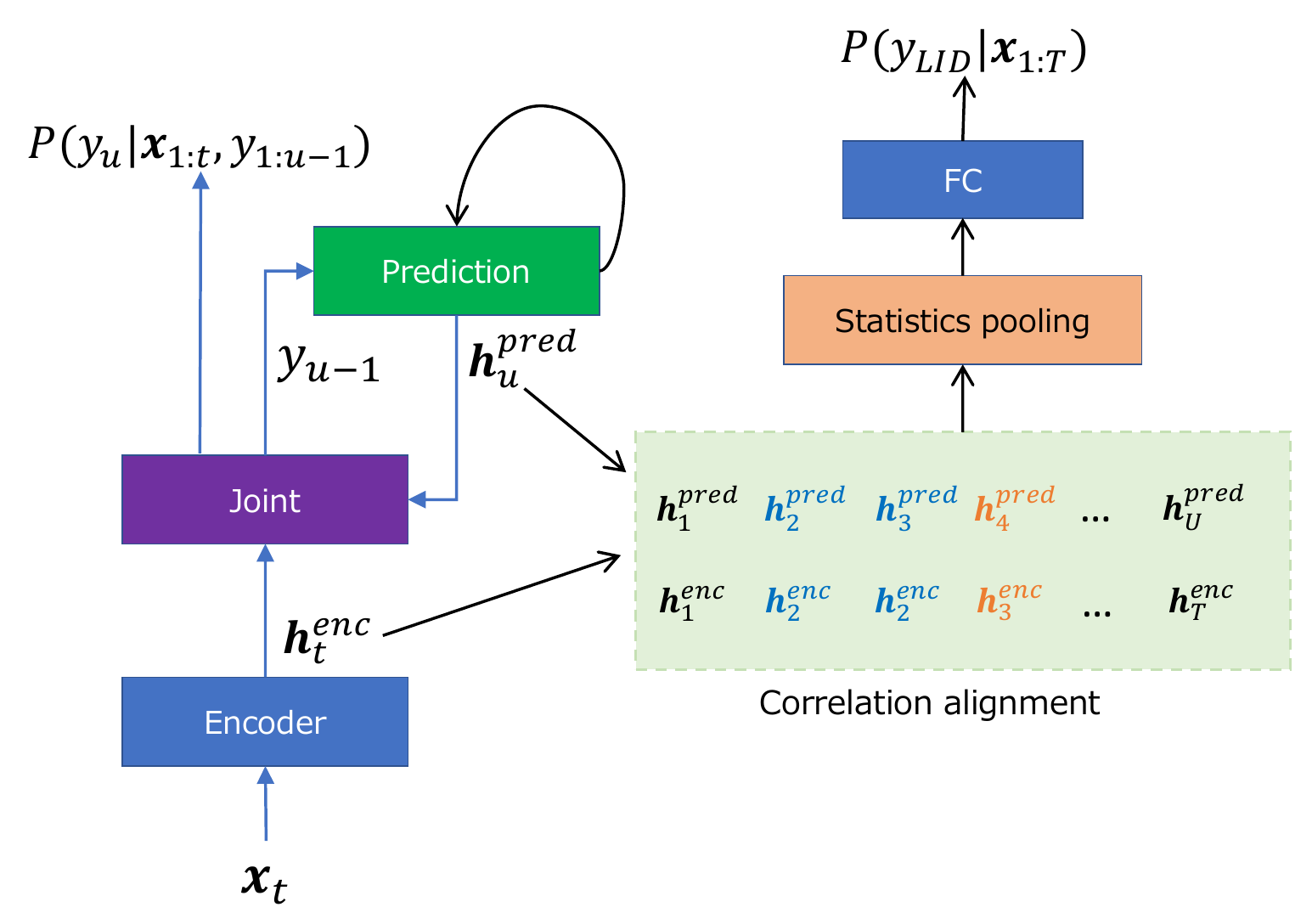}\\
  \caption{Proposed transducer-based language embedding}\label{fig.proposed_method}
\end{figure}

Fig. \ref{fig.proposed_method} illustrates the proposed RNN-T-based language embedding extraction architecture.
The front-end architecture consists of three blocks, i.e., an RNN-T block, a correlation alignment operation, and an utterance-level statistics pooling block.
The RNN-T block is used for frame-based feature extraction which consists of an encoder, prediction, and joint network. Then, the outputs of the encoder and prediction networks are processed with the correlation alignment.
Finally, the statistics pooling block is used to obtain the utterance-level language embedding features.

\subsubsection{Correlation alignment}
The RNN-T block maps an input sequence to an output sequence of arbitrary length. That is, for each input $\mathbf{x}_t$, the model will keep outputting symbols until a blank symbol $\phi$ is encountered.
If a blank symbol is encountered, the RNN-T reads in the next input $\mathbf{x}_{t+1}$ and repeats the output process until all input sequences have been processed.
Let us set the max number of output symbols to $\tau$. In most cases, the length is the difference between the audio inputs and the outputs of joint and prediction networks when $\tau > 1$.
Therefore the length of $\mathbf{h}^{enc}$ and $\mathbf{h}^{pred}$ is also different.
If fusion was done on $\mathbf{h}^{enc}$ and  $\mathbf{h}^{pred}$ for language embedding extraction, an alignment is needed to match them.

For an encoder outputs sequence of $\{ \mathbf{h}^{enc}_1, \mathbf{h}^{enc}_2,\mathbf{h}^{enc}_3, ... \}$ and the joint network outputs with $\{ \phi, h, e, \phi, ... \}$, the corresponding output sequence of the prediction network is described as $\{ \mathbf{h}^{pred}_1, \mathbf{h}^{pred}_2, \mathbf{h}^{pred}_3, \mathbf{h}^{pred}_4, ... \}$.
Although the blank symbol $\phi$ is not be optimized in the prediction network, the corresponding acoustic representation, i.e., $\mathbf{h}^{enc}_t$, still needs to be kept for language embedding extraction when the joint network emits blank symbols.
Therefore, we align the outputs of the encoder and prediction to $\{ [ \mathbf{h}^{enc}_1, \mathbf{h}^{pred}_1 ], [ \mathbf{h}^{enc}_2,  \mathbf{h}^{pred}_2 ], [ \mathbf{h}^{enc}_2,  \mathbf{h}^{pred}_3 ], [ \mathbf{h}^{enc}_3,  \mathbf{h}^{pred}_4 ], ... \}$.
When $\tau$ is set to $1$, the alignment mapping is simply done by collecting the two vectors one by one.

\subsubsection{Early additive fusion}

With the aligned outputs of the encoder and prediction networks, we can merge them as
\begin{equation}\label{eq.constraint}
  \mathbf{h_v} = \mathbf{h}^{enc}_t+\lambda \mathbf{h}^{pred}_u,
\end{equation}
where $\lambda$ is a weight coefficient to balance the contribution of the acoustic and linguistic representations.
The concatenation-based fusion can also be used.
To keep the operation the same as the pre-trained RNN-T model, in this work, we use the early additive fusion by setting $\lambda$ to 1.

\subsubsection{Language embedding}
Our language embedding extraction method is similar to x-vector \cite{snyer2018xvector4LID, Synder2018}. The x-vector representation model consists of three modules: a frame-level feature extractor, a statistics pooling layer, and utterance-level representation layers.
For frame-level features, time-delay neural network (TDNN) \cite{snyer2018xvector4LID} and convolutional neural network \cite{Li2017DeepSA} are widely used to prepare them. In this work, we use the fusion outputs $\mathbf{h_v}$ as the input frame-level feature.
A statistics pooling layer converts the frame-level feature $\mathbf{h}_v$
into a fixed-dimensional vector $\mathbf{h}_{p}$ by concatenating the mean $\bm{\mu}$ and standard deviation $\bm{\sigma}$ of them. Finally, fully-connected hidden layers are used to process the utterance-level representation and a softmax layer is used as the output with each of its output nodes corresponding to one language ID. The operations can be described as follows:
\begin{equation}\label{eq:xvector.01}
    \begin{split}
    \mathbf{h}_{p} = Pooling(\mathbf{h}_{1:V}),
    \end{split}
\end{equation}
\begin{equation}\label{eq:xvector.02}
    \begin{split}
    \mathbf{h}_{fc} = FC_{2}(FC_1((\mathbf{h}_{p}))),
    \end{split}
\end{equation}
\begin{equation}\label{eq:xvector.03}
    \begin{split}
    P(y_{LID}|\mathbf{x}_{1:T}) = softmax(\mathbf{h}_{fc}).
    \end{split}
\end{equation}
The output before activations of the first fully-connected layer after the statistics pooling layer is extracted as the language embedding representation.

\subsubsection{Late statistic pooling-based fusion}
In the early additive fusion, the outputs of the encoder and prediction are merged directly.
We also evaluated the late statistic pooling-based fusion method by merging the two representations after statistic pooling that can be described as
\begin{equation}\label{eq.fusion2.1}
    \begin{split}
      \mathbf{h}_{fc} = \mathbf{h}_{fc}^{enc}+\alpha \mathbf{h}_{fc}^{pred},
    \end{split}
\end{equation}
where $\alpha$ is a weight coefficient to balance the contribution of the acoustic and linguistic representations, and $\mathbf{h}_{fc}^{enc}$ and $\mathbf{h}_{fc}^{pred}$ are the outputs of Eqs. \ref{eq:xvector.01} and \ref{eq:xvector.02} with inputs of $\mathbf{h}_{1:T}^{enc}$ and $\mathbf{h}_{1:U}^{pred}$, respectively.
The output of Eq. \ref{eq.fusion2.1} is used for extracting the language embedding representation.

\subsection{Back-end classifier modeling}

For classification, a back-end classifier model is utilized. Although the front-end system is trained based on supervised learning with language IDs as labels, the feature from the embedding layer also encodes acoustic factors other than only language variations. To remove non-language-specific factors, the linear discriminant analysis (LDA) based dimension reduction is applied on the language embedding.  After the LDA, together with a length normalization (L-norm) process, the feature is input to a discriminative classifier for LID. In this study, logistic regression (LR) is applied. Considering the complex distribution of language clusters, a mixture of LR is adopted.

\section{Experiments and results}
\subsection{Dataset and evaluation metrics}
A large-scale multilingual LibriSpeech (MLS) dataset \cite{Pratap2020MLSAL} was used to evaluate the proposed method.
The MLS dataset consists of 8 languages, i.e., English, German, Dutch, French, Spanish, Italian, Portuguese, and Polish.
The original MLS includes 44.5K hours of English, and a total of 6K hours spread over seven other languages.
To reduce the influence of imbalance in the training data of different languages, we collected a maximum of 50,000 utterances for each language.
We made a validation set by randomly selecting 5,000 utterances, and keeping the remaining data as the training set.
For the test dataset, besides the official test dataset, we also prepared shorter utterances of test data by cutting fixed-length utterances of 1.0, 2.0, and 3.0 seconds from the original full-length test set.

To evaluate the performance of cross-domain data, we prepared a Youtube data-based dataset of VoxLingua107 \cite{valk2021vox107slt} by selecting the data of the target eight languages.
Because the official evaluation dataset doesn't consist of data from all the target eight languages, we prepared a new test dataset based on the original training data.
To reduce the possibility that the same speaker belongs to both the training and test datasets, we prepared the test dataset by randomly selecting 20 videos for each language, and keeping the remaining videos as the VoxLingua107 training dataset.
A model trained with the VoxLingua107 training dataset was used for preparing the oracle results.
The final test dataset consists of 5,437 utterances in total and the range of utterance numbers for each language is between 344 to 1,259. Similar to the MLS dataset, shorter duration datasets of 1.0, 2.0, and 3.0 seconds were also prepared.

The average performance cost (Cavg) described in \cite{AP19} is used as the evaluation metric.

\subsection{Experimental setup}
Several baseline systems were built with reference to LID and speaker recognition tasks \cite{AP20, snyder2019etdnn, Desplanques2020ECAPATDNNEC}.
The LID system contains a language embedding front-end and a logistic regression-based back-end. For the frame-level feature extraction network of front-end, we evaluated the extended TDNN (ETDNN) \cite{snyder2019etdnn} and ECAPA \cite{Desplanques2020ECAPATDNNEC} configurations. The ETDNN is widely used for both LID and speaker recognition tasks and the ECAPA obtained state-of-the-art performance on speaker recognition tasks \cite{Desplanques2020ECAPATDNNEC}.
There are two fully-connected layers with 512 neurons after the statistic pooling layer.
The input acoustic feature were 30-dimensional MFCCs (MFCC30) extracted from 36 Mel band bins-based spectrum. And the MFCCs are extracted with 25 ms frame length and 10 ms frameshift. Besides the MFCC feature, an 80-dimensional log Mel-based filter-bank feature (FBANK80) was also used in this work.
For improving the robustness, the SpecAugment technique was applied during the model training \cite{Park2019SpecAugment}.
The final extracted language embedding was with 512 dimensions. After the embedding was extracted, LDA-based dimension reduction and length normalization were applied before the logistic regression classier was applied. The dimension of LDA was set to 7.

For the proposed method, we used the conformer (large) network for the RNN-T encoder and one LSTM layer for the prediction network by referencing to \cite{conformer2020}.
The conformer encoder consists of 17 conformer blocks with an encoder dimensional of 512, attention heads set to 8, and convolutional kernel size set to 32. Subsampling with factor 4 was processed before the conformer layers.
The prediction network consists of 640 LSTM memory cells.
The outputs of the encoder and prediction network were transformed to 640 dimensions vectors with a linear transform layer of the joint network. The neurons of the final fully-collected layer of the joint network were 640.
We initialized the RNN-T model with the NeMo STT En Conformer-Transducer Large model\footnote{https://catalog.ngc.nvidia.com/models} trained with the large-scale NeMo ASRSET 2.0 dataset.

For baseline model training, we used the Adam algorithm with an initial learning rate of $0.001$. A warm restart learning rate schedular was used to adjust the learning rate during training \cite{Loshchilov2016Warmup}. The mini-batch size was 512 for ETDNN, and 128 for ECAPA.  The number of training epochs was set to 21.
Transducer-based methods were optimized using the Adam optimizer and batch size of 64, and an initial learning rate of $0.0001$, which is decayed by 10 when validation loss plateaus. The minimum learning rate was set to 1e-8. The baseline and proposed method systems were implemented based on the ASV-Subtools \cite{tong2021asv} and NVIDIA NeMo\cite{kuchaiev2019nemo} toolkits.

\subsection{Investigation and experimental results}
Table \ref{result.inDomain} shows the baseline results and the proposed method on the MLS datasets.
For baseline systems, we evaluated configurations of ETDNN and ECAPA with MFCC30 and FBANK80.
From the results, we can see that the ETDNN obtained better results than the ECAPA on our LID tasks. Although the ECAPA obtained state-of-the-art performance on speaker recognition tasks, we could not observe better performance on the LID datasets.

The transducer-based language embedding (TransVector) was trained only on FBANK80 since the RNN-T model was pre-trained with only the FBANK80 feature.
To better understand the proposed method, we conducted experiments to investigate the influence of the outputs of encoder, prediction, and joint networks, the max number of output symbols, the contribution balance of the acoustic and linguistic representation, and performance on cross-domain data.

\begin{table}[tb]
\centering
\caption{Experimental results (Cavg) of baseline systems and proposed method on in-domain MLS test datasets.}
\setlength{\tabcolsep}{0.2em}
\begin{tabular}{|l|c|c|c|c|c|c|} \hline
Models               & 1s     & 2s     & 3s     & Full   \\ \hline \hline
ETDNN-MFCC30           & \textbf{0.1561} & \textbf{0.0671} & \textbf{0.0431} & 0.0092 \\ \hline
ETDNN-FBANK80          & 0.1551 & 0.0743 & 0.0460 & \textbf{0.0085} \\ \hline
ECAPA-TDNN-MFCC30           & 0.1826 & 0.1259 & 0.0986 & 0.0503 \\ \hline
ECAPA-TDNN-FBANK80          & 0.1862 & 0.1109 & 0.0810 & 0.0354 \\ \hline \hline
Conformer encoder-based       & 0.1176 & 0.0497 & 0.0281 & 0.0091 \\ \hline
Prediction-based       & 0.3642 & 0.2958 & 0.2519 &	0.0977 \\ \hline
TransVector(Joint)     & 0.0973 & 0.0327 & 0.0165 &	0.0033 \\ \hline
TransVector(Early $\tau:3$)     & \textbf{0.0886} &	0.0341 & 0.0180 & 0.0054 \\ \hline
TransVector(Late $\alpha:0.1,\tau:3$) & 0.0978 & 0.0319& 0.0155 & 0.0033 \\ \hline
TransVector(Late $\alpha:0.3,\tau:3$) & 0.0925 & \textbf{0.0307} & \textbf{0.0143} & \textbf{0.0026} \\ \hline
TransVector(Late $\alpha:0.5,\tau:3$) & 0.1034 & 0.0350 & 0.0171 & 0.0037 \\ \hline \hline

TransVector(Early $\tau:1$)  & 0.0970 & 0.0337 & 0.0172 & 0.0039 \\ \hline
TransVector(Late $\alpha:0.3,\tau:1$) & 0.0922 & 0.0297 & 0.0145 & 0.0031 \\ \hline
\end{tabular}
\vspace{-3mm}
\label{result.inDomain}
\end{table}

\textbf{Outputs of different RNN-T network:}
We investigated the LID performance by comparing language embedding built with the outputs of the encoder, prediction, and joint network.
The parameters of prediction and joint networks were not updated when the model was trained using the output based on the encoder, i.e., $\mathbf{h}^{enc}$. For comparison, we also fixed the parameters of the encoder and joint networks when the model was trained with the output of the prediction network, i.e., $\mathbf{h}^{pred}$.
Both the encoder and prediction networks were optimized when using the outputs of the joint network.
The model trained with $\mathbf{h}^{enc}$ can be considered using the acoustic or implicit linguistic representation of the ASR model.
The model using the output of the joint network can be considered using both acoustic and linguistic information.
The results showed that compared with the ETDNN system, the model trained with the encoder-based feature significantly improved the performance.
Using only the prediction-based linguistic information cannot improve the performance, this may be because the update of the parameters of the encoder was restricted.
The model trained based on the output of the joint network performed better than using the output of the conformer encoder network.

\textbf{Effect of early and late fusion:}
Results in Tables. \ref{result.inDomain} and \ref{result.outDomain} showed that the early fusion method outperformed the method using the encoder-based acoustic features and achieved comparable or better performance compared with the joint-based methods.
The late statistic pooling fusion could further improve the performance, especially on the relative longer utterances, i.e., duration longer than 2s.

\textbf{Effect of transducer symbol output number :}
We investigated the influence of the max number output symbol $\tau$ by comparing the values of 1 and 3.
From the results in Tables \ref{result.inDomain} and \ref{result.outDomain}, we can see the max number setting has no noticeable effect on in-domain data. However, on the cross-domain datasets, the performance was improved by setting it to 3.

\textbf{Performance on cross-domain dataset:} Table. \ref{result.outDomain} listed the results on the cross-domain datasets, i.e., using MLS data for model and classifier building and evaluated on the Voxlinga107 test data.
The results of the ETDNN system trained with the MFCC30 feature on MLS training data are listed for comparison.
We also prepared ETDNN oracle results obtained with the system trained with the Voxlingua107 training dataset.
From the results, we can see that the proposed method trained with acoustic and explicit linguistic feature significantly improved the performance of the cross-domain dataset.

\begin{table}[tb]
\centering
\caption{Experimental results (Cavg) of on cross-domain Voxlingua107 test datasets.}
\setlength{\tabcolsep}{0.2em}
\begin{tabular}{|l|c|c|c|c|c|c|} \hline
Models                & 1s     & 2s     & 3s     & Full   \\ \hline \hline
ETDNN-MFCC30                   & 0.2450 & 0.1774 & 0.1478 & 0.1197 \\ \hline
ETDNN-MFCC30 (Oracle)           & 0.1176 & 0.0507 & 0.0278 & 0.0041 \\ \hline \hline

Conformer encoder-based        & 0.2112 & 0.1297 & 0.1084 & 0.0832 \\ \hline
TransVector(Joint)      & 0.1710 & 0.1056 & 0.0827 & 0.0700 \\ \hline

TransVector(Early $\tau:3$)             & 0.1653 & 0.1019 & 0.0840 & 0.0655 \\ \hline
TransVector(Late $\alpha:0.3,\tau:3$)   & \textbf{0.1635} & \textbf{0.0931} & \textbf{0.0716} & \textbf{0.0544} \\ \hline
TransVector(Late $\alpha:0.3,\tau:1$)   & 0.1672 & 0.0956 & 0.0738 & 0.0574 \\ \hline
\end{tabular}
\vspace{-3mm}
\label{result.outDomain}
\end{table}

\subsection{Discussion}

Similar to previous works, our experimental results also showed that using a pre-trained ASR model could improve the performance on LID tasks. It is mainly because the pre-trained ASR encodes the phonetic or implicit linguistic information which is beneficial for the LID task.
Our experimental results also showed that comparing models using only the encoder-based acoustic feature, models trained with both the acoustic and explicit linguistic representation could significantly improve the performance.
The investigations on the early- and late-fusion methods showed that a tunable parameter was an effective strategy for effectively utilizing the acoustic and linguistic features.
To keep the operation the same as that of the RNN-T joint network, the early additive fusion was not investigated in depth in this work. Further investigation of the fusion method is one of our future works.

Compared with the ETDNN baseline system, the late fusion method obtained 35\% to 71\% relative improvement on the in-domain dataset, and 33\% to 54\% on cross-domain datasets.
Compared with the conformer encoder-based method, it obtained 12\% to 59\% and 16\% to 24\% relative improvement on in-domain and cross-domain datasets, respectively.
The experimental results showed that the proposed TransVector method is an effective method for LID tasks.

\section{Conclusions}
In this paper, we propose a novel transducer-based language embedding method that integrates the RNN-T model into a language embedding extraction framework. Benefiting from the advantages of the RNN-T's acoustic and linguistic representation, the proposed method can encode both acoustic and explicit linguistic features for building robust LID systems.
To better utilize the acoustic and linguistic information, we evaluated early hidden layer-based fusion and late statistic pooling-based fusion methods.
The experimental results showed the effectiveness of the proposed method on both the in-domain dataset and the cross-domain dataset for LID tasks.

\section{Acknowledgements}
This work is partially supported by JSPS KAKENHI No.19K12035 and No.21K17776.

\newpage
\bibliographystyle{IEEEtran}
\bibliography{mybib}

\end{document}